\documentclass[preprint]{article}

\PassOptionsToPackage{numbers, compress}{natbib}
\usepackage{main}

\usepackage[utf8]{inputenc} 
\usepackage[T1]{fontenc}    
\usepackage{hyperref}       
\usepackage{url}            
\usepackage{booktabs}       
\usepackage{amsfonts}       
\usepackage{nicefrac}       
\usepackage{microtype}      
\usepackage{xcolor}         
\usepackage{graphicx}
\usepackage{float}
\usepackage{subfig}
\usepackage{longtable}
\usepackage{bbding}
\usepackage{longtable}

\title{Large Language Model as a Universal Clinical Multi-task Decoder}

\author{%
    Yujiang Wu$^{1}$\footnotemark[1]  \footnotemark[2] , Hongjian Song$^{2}$\footnotemark[1] \footnotemark[2] , Jiawen Zhang$^{3}$\footnotemark[2],\\
    \textbf{Xumeng Wen$^{4}$, Shun Zheng$^{4}$, Jiang Bian$^{4}$} \\
    $^{1}$Carnegie Mellon University, $^{2}$South China University of Technology, \\
    $^{3}$HKUST (GZ), $^{4}$Microsoft Research Asia\\
  \texttt{yujiangw@andrew.cmu.edu}, \texttt{shjsonghongjian@gmail.com} \\
  \texttt{jzhang302@connect.hkust-gz.edu.cn} \\
  \texttt{\{xumengwen, shun.zheng, jiang.bian\}@microsoft.com} \\
}

\begin{document}

\renewcommand{\thefootnote}{\fnsymbol{footnote}}
\footnotetext[1]{The first two authors contributed equally.}
\footnotetext[2]{This work was done during the internship at Microsoft Research Asia, Beijing, China}

\maketitle

\begin{abstract}
\label{abstract}
The development of effective machine learning methodologies for enhancing the efficiency and accuracy of clinical systems is crucial. Despite significant research efforts, managing a plethora of diversified clinical tasks and adapting to emerging new tasks remain significant challenges. This paper presents a novel paradigm that employs a pre-trained large language model as a universal clinical multi-task decoder. This approach leverages the flexibility and diversity of language expressions to handle task topic variations and associated arguments. The introduction of a new task simply requires the addition of a new instruction template. We validate this framework across hundreds of tasks, demonstrating its robustness in facilitating multi-task predictions, performing on par with traditional multi-task learning and single-task learning approaches. Moreover, it shows exceptional adaptability to new tasks, with impressive zero-shot performance in some instances and superior data efficiency in few-shot scenarios. This novel approach offers a unified solution to manage a wide array of new and emerging tasks in clinical applications.
\end{abstract}

\section{Introduction}
\label{Introduction}

The development of effective machine learning methodologies for enhancing the proficiency and precision of clinical systems is of utmost importance. These advancements are vital for improving diagnostic, prognostic, and decision-making processes and have attracted significant research interest in recent years~\cite{harutyunyan2019multitask,heo2019machine,johnson2016mimic,li2018tatc,mcdermott2021comprehensive,muralitharan2021machine,nezhad2019deep}.

A substantial portion of existing research~\cite{che2018recurrent,luo2020hitanet,shukla2021multitime,tipirneni2022strats,zhang2023warpformer} is dedicated to understanding how to derive effective representations from heterogeneous and irregularly sampled clinical signals, such as heart rates, glucose levels, and sodium concentrations. The goal is to inform downstream tasks such as early warning, outcome prediction, and treatment recommendations. These studies have proficiently addressed the challenges of representation learning for clinical time series, particularly the irregular sampling intervals and vast discrepancies between different signals~\cite{zhang2023warpformer}.

While input modeling presents its challenges, another characteristic of AI-assisted clinical systems is the necessity to manage a plethora of diversified clinical tasks~\cite{mcdermott2021comprehensive}. These tasks span different topics and purposes, such as mortality prediction (MOR), decompensation (DEC), length of stay (LOS), phenotype classification (Phenotype), and next timepoint Will be measured (WBM). Even within a specific task category, there are numerous variations due to different choices of prediction windows, classification taxonomy, and label choices. This clinical task system is also continually evolving with the introduction of new measurements and phenotypes.

The sheer volume and continual evolution of clinical tasks pose a significant challenge for machine learning methodologies. Building a single model for each task, or single-task learning (STL), can lead to an overwhelming amount of model checkpoints to manage, which is inefficient. Multi-task learning (MTL) approaches offer some relief; however, their effectiveness in clinical scenarios is still under debate due to potential performance deteriorations in some tasks~\cite{harutyunyan2019multitask}. Suggestions for dividing tasks into different groups for MTL have been put forth~\cite{song2022mtgnet}. Even though, the introduction of a new task necessitates a complicated learning process, considering how to train and deploy the model for this task in conjunction with existing tasks.

To navigate the challenge of managing massive clinical tasks and flexibly adapting to emerging new tasks, this paper proposes a novel paradigm. We utilize a pre-trained large language model as a universal clinical multi-task decoder, leveraging the flexibility and diversity of language expressions to handle task topic variations and associated arguments. The introduction of a new task simply requires the addition of a new instruction template. Considering the heterogeneity and irregularity of clinical data, we opt to leverage existing clinical representation learning approaches, such as Warpformer. We develop an adapter to bridge the output representation of Warpformer and the input of the large language model, enabling gradient backpropagation from the model output into Warpformer to learn universal representation for all kinds of tasks.

In our experiments, we have validated the effectiveness of the proposed ClinTS-LLM framework across hundreds of tasks. This framework has proven to robustly facilitate multi-task predictions, performing on par with traditional MTL and STL approaches. A unique advantage of this framework is its adaptability to new tasks. We have evaluated its zero-shot and few-shot behaviors on select holdout tasks, observing exceptional zero-shot performance in some instances. Moreover, its few-shot learning demonstrates superior data efficiency compared to traditional methods.

In summary, our contributions are as follows:
\vspace{-6pt}
\begin{itemize}
    \item In response to the multi-task challenge in clinical applications, we have pioneered a novel paradigm that integrates traditional clinical data representation learning with a modern pretrained LLM. The resulting model is capable of managing a vast array of new and emerging tasks in a unified manner.
    \item We have conducted extensive experiments to evaluate this framework. Specifically, we have demonstrated that its MTL performance rivals that of traditional MTL and STL methods. Uniquely, it excels in its ability to transfer to new tasks, showcasing impressive zero-shot transfer in some instances and enhanced data efficiency in few-shot scenarios.
\end{itemize}
\vspace{-8pt}

\begin{figure*}[t]
  \centering
    \subfloat[Example clinical data.]{
    \label{fig:icu_stay}
    \includegraphics[width=0.45\linewidth]{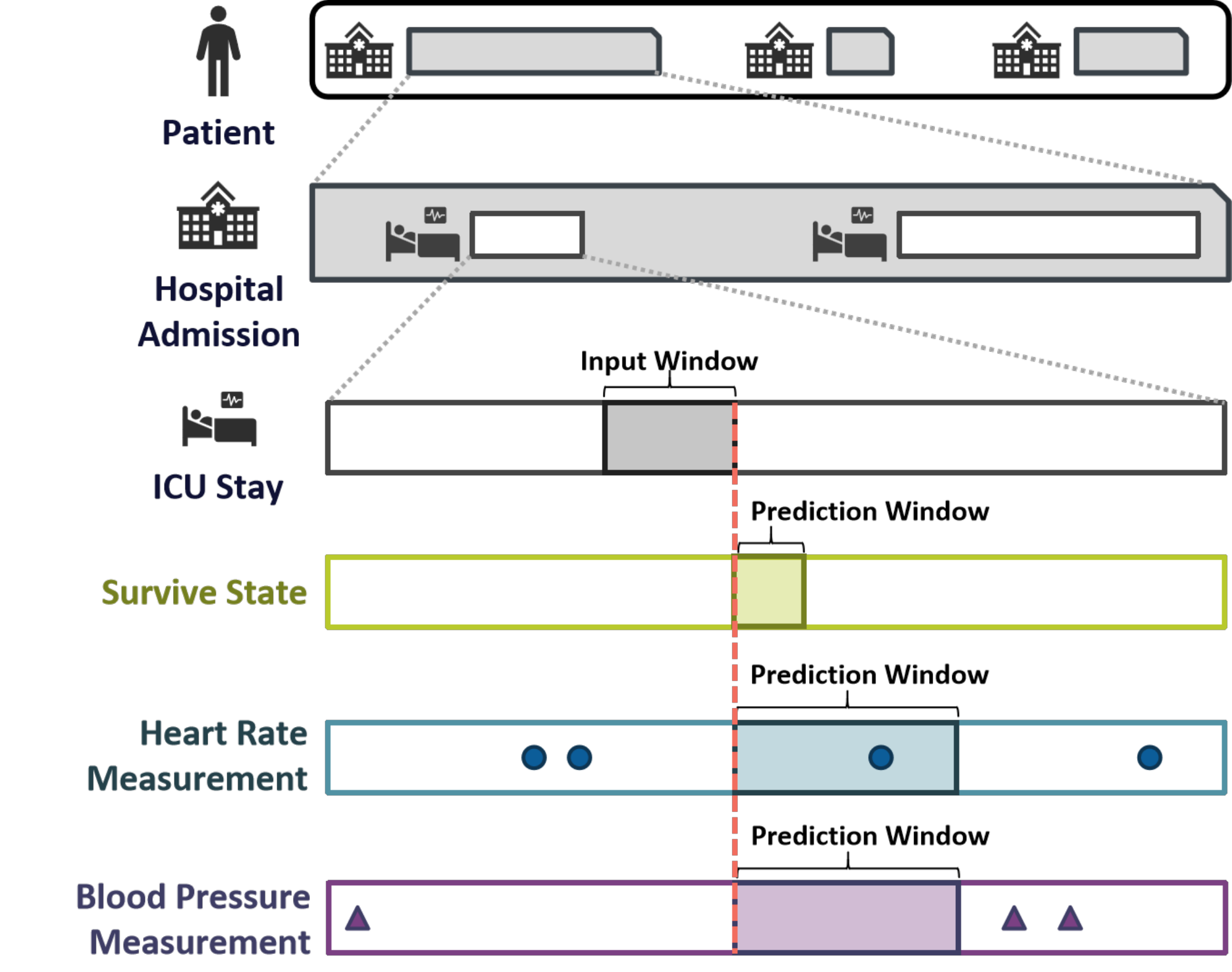}}
    \hspace{1em}
    \subfloat[Task organization.]{
    \label{fig:tasks}
    \includegraphics[width=0.5\linewidth]{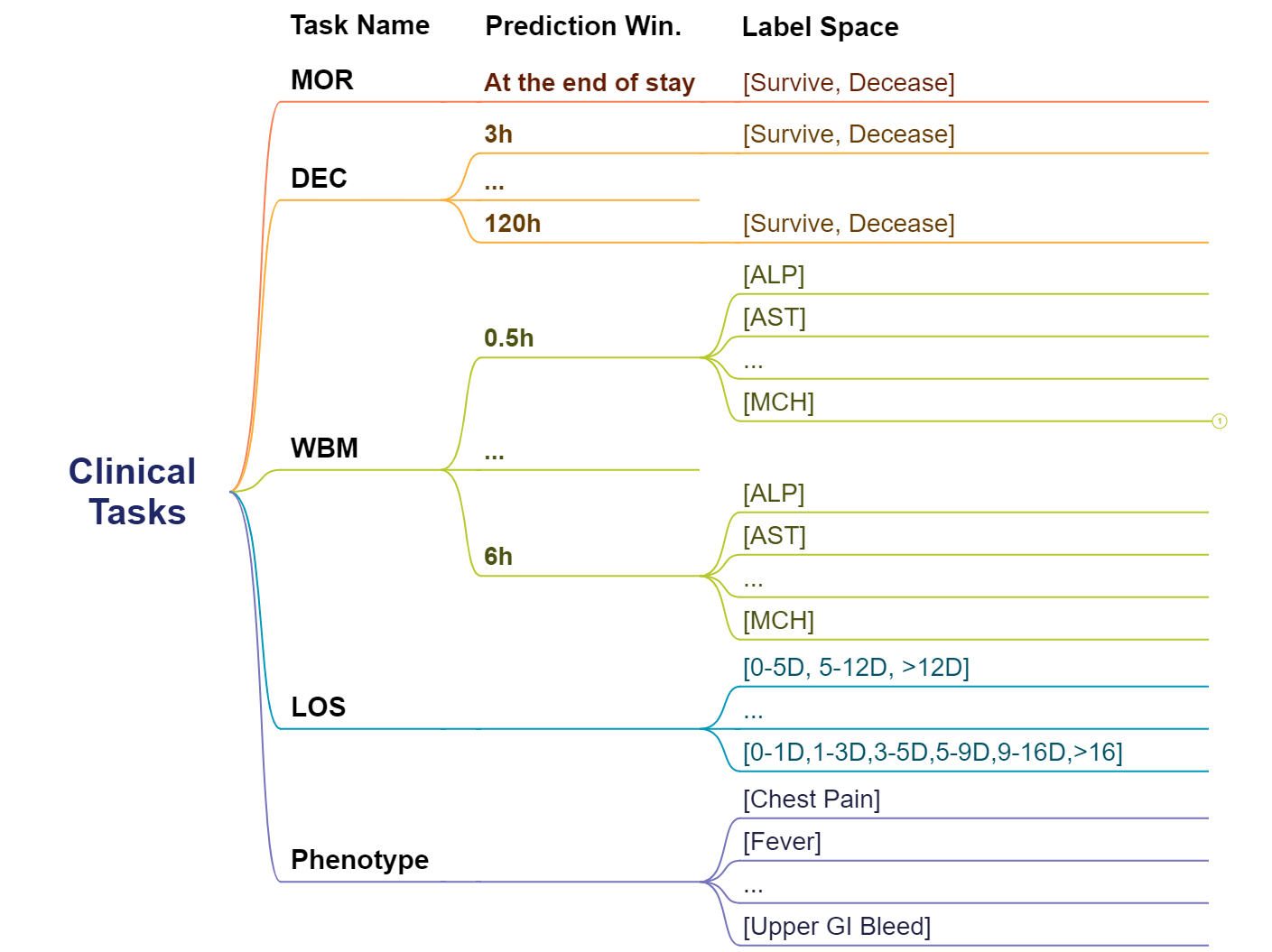}}
    \caption{An overview of clinical data collection and task organization.}
    \vspace{-15pt}
\end{figure*}





\section{Related Work}
\label{sec:rel_work}

\emph{\textbf{Clinical Time-series Modeling.}}
The modeling of clinical time-series data is a critical research area. Recent years have witnessed a surge in benchmarks for clinical data~\cite{harutyunyan2019multitask,jarrett2021clairvoyance,McDermott2021chil,wang2020mimic, yeche2021HiRID-ICU}, promoting a more standardized approach to model evaluation. However, these benchmarks often cover a narrow range of dataset settings. Various methods have been developed to enhance prediction accuracy and efficiency in healthcare~\cite{luo2020hitanet,MaGWZWRTGM20,tipirneni2022strats,xu2018raim, MUFASA, zhang2023warpformer,ZhangQLLCGL21}, but they are usually confined to specific tasks with predetermined configurations.



\emph{\textbf{Multitask Learning in Healthcare.}}
Multitask learning, which leverages commonalities and variations across tasks, aims to enhance model generalization and performance for individual tasks~\cite{baxter2000model,Caruana93multitask, caruana1997multitask, zhang2021survey}. 
This approach is particularly relevant in healthcare, where clinical objectives are diverse yet interconnected. Shared representations from multitask learning have been shown to improve outcomes by exploiting inter-task correlations~\cite{ daniels2021multitask, ding2018effectiveness,harutyunyan2019multitask, lam2022multitask, suresh2018learning}.
However, the numerous possible task configurations in healthcare often challenge traditional multitask learning models. Minor alterations in task setup, such as a shift in the prediction window, can drastically reduce the efficacy of a learned task decoder, requiring additional labeled data for retraining. This limitation highlights the need for a more adaptable, universal task decoder, particularly one effective in zero-shot settings. Large Language Models, with their flexibility and ability to process diverse data formats and scenarios, present a promising solution for multitask learning challenges in complex clinical environments.

\emph{\textbf{Large Language Models for Time Series}}. Most research on LLMs in the time series scenarios focuses on forecasting tasks. These studies emphasize that LLMs pre-trained across a diverse range of domains (such as traffic, finance, healthcare, etc.) can achieve strong zero-shot performance on new datasets \cite{cao2024tempo,chang2023llm4ts, das2023decoder,  jin2023time, gruver2023llmtime, zhou2023one}. However, these models are not specifically designed for healthcare applications.
In contrast, there is limited research on using Large Language Models (LLMs) for time series classification tasks. Studies such as \cite{li2024frozen,liu2023large} have applied LLMs for zero-shot classification in healthcare, but they focus on single-signal, simple tasks, which significantly differ from our approach in terms of task design and model architecture. In this paper, we explore the use of LLMs as a universal decoder, addressing multiple variables to perform complex clinical diagnostic tasks in both in-domain and zero-shot settings.

\section{Methodology}
\label{sec:method}
\subsection{Clinical Tasks}
\subsubsection{Data structure}
To facilitate a thorough explanation of our task selection and task division in the future, it is necessary to first introduce our data structure: In the MIMIC-III dataset, the largest unit is the patient, as $MIMIC - III \sim \left[ {Patien{t_1},Patien{t_2}, \cdots ,Patien{t_n}} \right]$. For each patient, there is at least one hospital admission (HADM), as $Patien{t_i} \sim \left[ {HAD{M_1},HAD{M_2}, \cdots ,HAD{M_n}} \right]$. For each hospital admission of a patient, they may require zero, one or multiple ICU treatments, as $HAD{M_i} \sim \left[ {ic{u_1},ic{u_2}, \cdots ,ic{u_n}} \right]$ (shown as Figure \ref{fig:icu_stay}). For each ICU treatment of a patient, various clinical events may occur, corresponding to task data and timestamps, as $ic{u_i} \sim \left( {t{s_i},dat{a_i}} \right)$, recorded in the MIMIC-III dataset. The ${dat{a_i}}$ represents the combination of input and all the corresponding labels of sample $i$, as $dat{a_i} = \left( {inpu{t_i},\left[ {labe{l_i^1}, \cdots ,labe{l_i^n}} \right]} \right)$. That is when constructing the input through sliding windows, we match each input with all the labels corresponding to it. The ${t{s_i}}$ represents the starting timestamp of ${dat{a_i}}$. Finally, to efficiently locate the data, we identify the positions of all inputs in the MIMIC-III dataset and link them to their corresponding labels. We achieve precise positioning of each input by maintaining an index for each one, in the form of: 
$\left( {Patien{t_i},HAD{M_i},ic{u_i},t{s_i}} \right)$.

\subsubsection{Task Selection}
\label{task select}

We have constructed a hierarchical and complex task system in the clinical field aimed at effectively treating and monitoring patients, as illustrated in Figure \ref{fig:tasks}. Specifically, the system include five groups of tasks: In-hospital Mortality (Mor), Decompensation (Decom), Length of Stay (LOS), Next Timepoint Will Be Measured (WBM), and Phenotype. These tasks vary in complexity, with MOR having one level, WBM comprising three levels, and the other tasks having two levels. As described in Section \ref{basis dividing dataset}, we use a time-based approach to divide the training, validation, and test sets. All tasks share the same input: a 24-hour sliding window on multiple time series signals. The constructing method and the label of each group's tasks are as follows (more details, as well as the differences between the clinical tasks of our work and the clinical tasks of other works, are shown in Section \ref{differences in clinical tasks}, and the detailed description of each specific task are shown in Table \ref{tab:all_task} in Section \ref{ap exp result}):


\emph{\textbf{In-hospital Mortality (MOR).}}
Label 1 indicates that the patient has died by the end of the current hospital admission, and 0 indicates that the patient has not died.

\emph{\textbf{Decompensation (Decom).}}
all the Decom tasks are constructed with a series of future windows from 3h to 120h, with 3h increments. 1 indicates the patient will die within the future window, and 0 indicates the patient will survive.

\emph{\textbf{Length Of Stay (LOS).}}
The different LOS tasks are built by categorizing the expected discharge dates within various date ranges. The number of categories ranges from 2 to 7, with detailed rules provided in Section \ref{differences in clinical tasks}.

\emph{\textbf{Phenotype.}}
The different phenotype tasks are constructed by selecting various sets of diseases. 1 indicates the patient has the disease by discharge, and 0 indicates they do not.

\emph{\textbf{Next Timepoint Will Be Measured (WBM).}}
Predicts required clinical interventions within future windows from 0.5h to 6h, with 0.5h increments. 1 indicates an intervention is needed within the window, and 0 indicates it is not.

\subsubsection{Task Transfer.}
\label{task transfer}
We categorize clinical tasks in the system into two groups: original tasks and transfer tasks. For original tasks, we require a substantial amount of data and employ Multitask Learning (MTL) to train our LLM+Warpformer model. Once the model has been adequately trained on the original tasks and has shown generalization capabilities across various clinical tasks, we then evaluate its transfer ability with zero-shot or few-shot methods on the transfer tasks. These tasks are derived from the original tasks but differ in several aspects. Below, we detail the different types of transfer methods employed.

\emph{\textbf{Parameter transfer.}}
Parameter transfer modifies parameters of the original clinical task. In this work, we focus on changing the future window size to evaluate the model's performance under different observation windows. This transfer includes in-domain and out-domain categories, varying in difficulty. For example, in the Decom task, original future windows range from "3h" to "117h". In transfer tasks, selecting windows within this range is in-domain, while a "120h" window is out-domain. Table \ref{tab:task} outlines the tasks that employ the parameter transfer method in constructing new tasks.

\emph{\textbf{Label transfer (Class partition).}}
Label transfer through class partition modifies the classification rules of multi-class tasks, specifically for Length of Stay (LOS) tasks. Similar to parameter transfer, it includes in-domain and out-domain categories. In LOS tasks, the original tasks have 2, 4, and 6 classifications. The transfer tasks with 3 and 5 classifications are considered an in-domain transfer, which falls within the original range, whereas a task with 7 classifications, which extends beyond the original range, is considered an out-domain transfer. The transfer details of the LOS task are shown in Table \ref{tab:task}.

\emph{\textbf{Label transfer (Label choice).}}
This task involves changing the label selection of multi-label tasks in the original tasks. In the original tasks, specific labels are chosen to construct multi-label tasks for model training. In transfer tasks, we use alternative labels that do not overlap with those selected in the original tasks to validate the model's transferability and generalization. The tasks that utilize label transfer (label choice) for constructing new tasks are detailed in Table \ref{tab:task}.

\begin{table*}
\centering
\small
\caption{\label{tab:task} Specifications for clinical tasks (\textit{BC}: binary classification, \textit{ML}: multi-label classification, \textit{MC}: multi-class classification). }
\begin{tabular}{lcccccc}
\toprule
\textbf{Task}  &  \textbf{Type} & \textbf{\# Subtask} & \textbf{Param.Transfer} & \textbf{Label.Transfer} \\
\midrule
In-hospital Mortality  & BC  &  1  & \XSolidBrush & \XSolidBrush  \\  
Decompensation      & BC  &  40 &   \CheckmarkBold & \XSolidBrush \\
Next Timepoint Will Be Measured    & ML   & 135  & \CheckmarkBold & \XSolidBrush \\
Length Of Stay  & MC  & 6   &  \XSolidBrush & \CheckmarkBold \\
Phenotype    & ML   & 14  &  \XSolidBrush & \CheckmarkBold \\
\bottomrule
\end{tabular}
\vspace{-5 pt}
\end{table*}

\subsection{Model Architecture}

\begin{figure*}[htbp] 
  \centering
    \subfloat[The design of propmt.]{
    \label{fig:prompt}
    \includegraphics[width=0.52\linewidth]{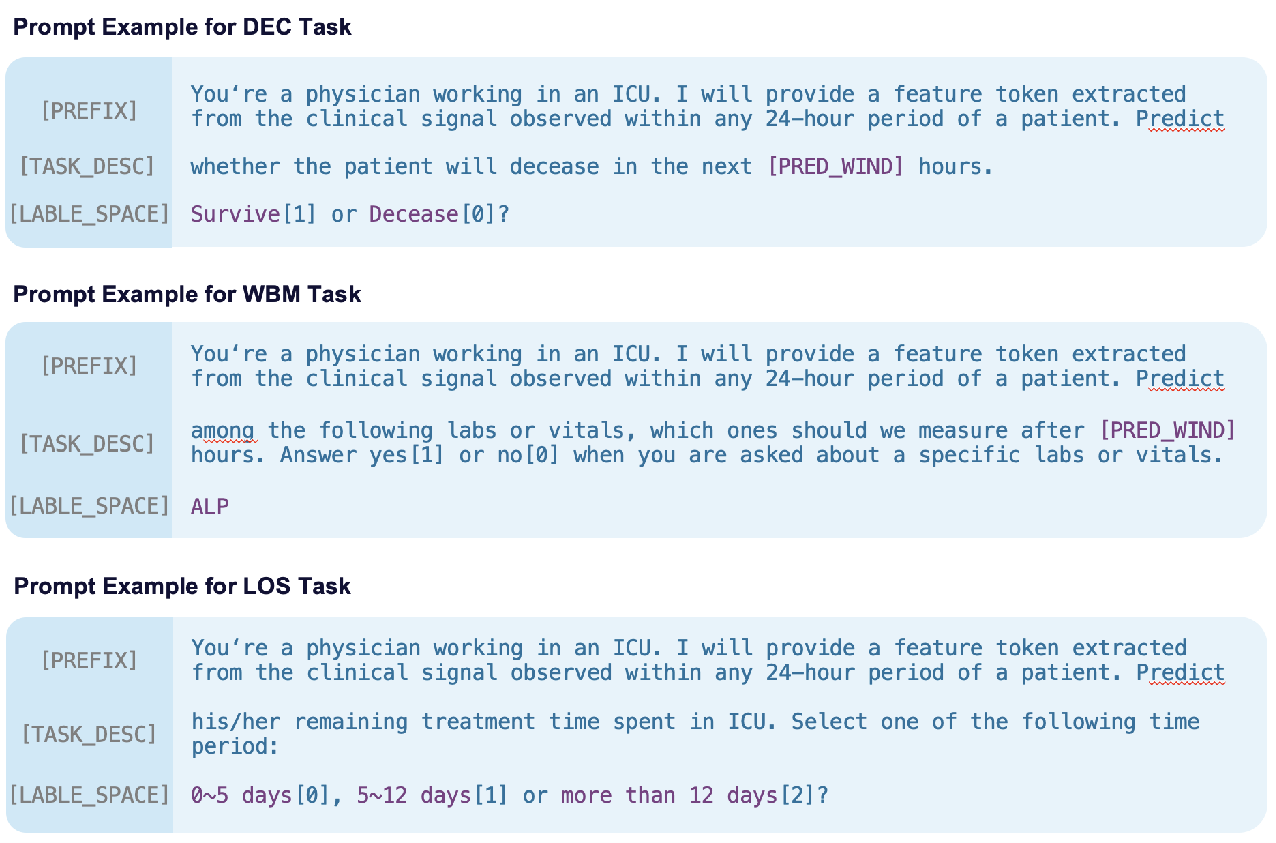}}
    \hspace{1em}
    \subfloat[The overview of our architecture.]{
    \label{fig:overview}
    \includegraphics[width=0.43\linewidth]{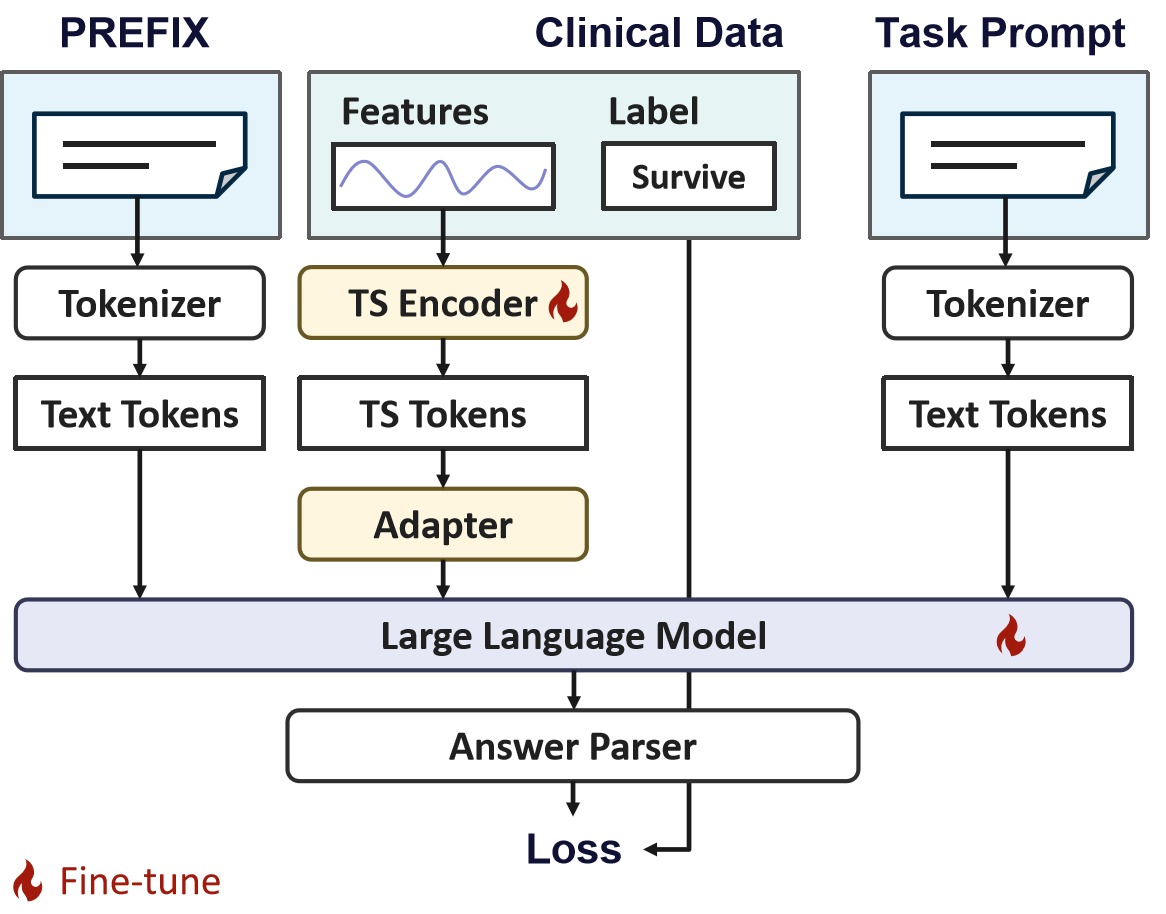}}
    
    \subfloat[The design of adapter.]{
    \label{fig:adapter}
    \includegraphics[width=0.9\linewidth]{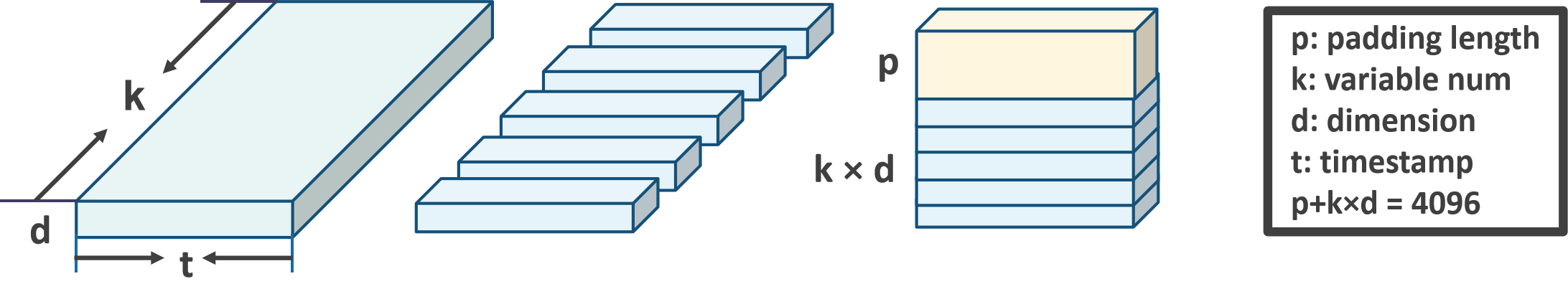}}
    \caption{The technical details for our model.}
    \label{fig:technical details}
\vspace{-10pt}
\end{figure*} 

\subsubsection{Overview}
The multivariate and irregular nature of clinical time-series tasks distinguishes them from typical time series scenario. Hence, instead of directly inserting values of variables in sequence into prompts, we employ the transformer backbone of Warpformer as the encoder and propose an effective adapter module to insert the features of clinical signals into natural language. The prefix context, task descriptions, embeddings of Warpformer, and ground truth, collectively form a prompt, which is then utilized for fine-tuning LLM with LoRA, as backbone of LLM frozen. Figure \ref{fig:technical details} illustrates the composition of the prompt, the overall architecture of the model, and the Adapter module.

\subsubsection{Clinical Time Series Encoder}
Warpformer\cite{zhang2023warpformer} is a Transformer model leverages $I$, a set of observations of multiple signals at irregular timestamps to generate prediction $P=\phi_d(\varphi_a(\psi_w(I))), I \in \mathbb{R}^{\widetilde{t} \times k}$, where $ \widetilde{t} $ denotes the union of timestamps of $k$ signals. $\psi_w$ is the Transformer backbone mapping $I$ into representation $x=\psi_w(I) \in \mathbb{R}^{t \times k \times d}$, where $t$ is the number of timestamps after warping and $d$ is the feature dimension of Warpformer. We discard the feature aggregation module $\varphi_a$ and the decoder $\phi_d$ in Warpformer. Instead, we feed $x$, the embedding  extracted by Warpformer's Transformer encoder into an Adapter module, described as follows, for further time-series embedding injection.

\subsubsection{adapter}
Given that the feature dimension of the LLM is significantly higher than that of Warpformer, we design an adapter module to bridge the embeddings from Warpformer to the LLM. Specifically, we unfold the time series features along the $t$ dimension, resulting in $\widetilde{x}=\rm{UnFold}(x) \in \mathbb{R}^{t \times (k*d)}$. We then pad $(k*d)$ to 4096, which corresponds to the feature dimension of Llama-7B, using zero vectors, as shown in Figure \ref{fig:adapter}. This prepares the time series embeddings for insertion into the prompt embeddings. Consequently, the time series embeddings occupy $t$ tokens.

\subsubsection{Prompt Design and Modal Fusion}
As illustrated in Figure \ref{fig:prompt}, our training prompt comprises a prefix, a task description, a time series embedding of length $t$, a label space to facilitate parsing the classification from language, and the ground truth label. These textual embedding are concatenated with the time series embedding produced by the adapter module. During inference, the ground truth label is excluded from the prompt.

\subsubsection{Multi-task Training and zero-shot Transfer}
Instead of fully tuning the LLM, we employ a Low-Rank Adapter (LoRA) \cite{hu2021lora} for efficiency. During training, we freeze the LLM and optimize the parameters of both LoRA and Warpformer, as in Figure \ref{fig:overview}, by minimizing the cross-entropy loss between the next token predicted by the LLM and the ground truth text at the corresponding position. At inference, we interpret the prediction by selecting the token at the label's corresponding position with the highest probability in the vocabulary. We train on five clinical prediction tasks, as introduced in Section \ref{task select}, using varied settings. We then evaluate the zero-shot performance on a held-out subset of settings, which the LLM has not encountered during training. The principles of the held-out setting for zero-shot transfer are detailed in Section \ref{task transfer}. More experiment details are shown in Section \ref{sec:exp_setting}.

\section{Experiments}
\label{sec:exp}
\subsection{Experimental Settings}
\subsubsection{Baselines Selection}
We roughly classify the existing methods that perform well on irregular time series clinical tasks into three categories. The first category is to introduce irregularity-sensitive updating mechanisms into the recurrent neural network, such as RNN-Mean (mean), RNN-Forward (forward), RNN-$\Delta t$ (delt), RNN-Decay (decay), and GRU-D ~\cite{che2018recurrent} (grud). The second category is to organize multiple types of irregular observation points into a long sequence of (time, type, value) tuples and model their interaction through attention mechanisms, such as ~\cite{horn2020set}. The third category is to introduce the encoding of features and maintain irregular features through warping in all subsequent operations, such as Warpformer ~\cite{zhang2023warpformer}. Considering that we need the LLM model to be a universal decoder on a large training task set, we can determine which model is most suitable as the backbone of LLM by comparing the effects of these three categories of methods on the training tasks using multi-task learning. The evaluation metrics of our experiments are shown in Section \ref{sec:evaulate_metric}.

\subsubsection{Implementation Details}
\textbf{Comprehensive Baseline Evaluation.} To avoid randomness when comparing the effects of various baselines, we selected three random seeds to initialize the model and used the final average and variance to reflect the experimental results. Given the unprecedented number of tasks included in our work, analyzing the performance of each baseline on individual tasks does not suffice to evaluate their overall performance across the entire training task system. Additionally, relying solely on mean and variance calculations lacks detailed evaluation. Therefore, we conducted a comprehensive assessment of the selected baselines' overall and detailed performance using scatter box plots, which present the effects of these baselines on all tasks and major classification tasks simultaneously.

\textbf{Evaluating Inference Performance.} To evaluate the effectiveness of our LLM+Warpformer architecture, we conducted experiments from two perspectives. First, we aimed to demonstrate that using LLM as a universal decoder for inference on the training task set is comparable to using multiple independent classifiers. Second, we compared the Multi-Task Learning (MTL) and Single-Task Learning (STL) approaches of Warpformer with LLM to show that training all tasks together does not significantly degrade performance, even with a large task set that includes hundreds of tasks influencing each other. Similar to the baseline comparison experiment, we presented the results using scatter box plots for both overall and task-specific evaluations. This approach ensures that LLM, as a universal decoder, performs comparably to MTL and STL of Warpformer across all tasks and does not exhibit significant weaknesses in any specific task category.

\textbf{Investigating Transferring Capabilities.} 
To explore the potential of LLM as a universal decoder for transferring to new domains, we selected reference models for comparison. We propose two training methods for the Multi-Task Learning (MTL) of Warpformer. Our Warpformer model is trained on multiple tasks across each major category (Decom, LOS, Phenotype, WBM) using sample sizes of 100, 200, 400, 800, and 1600. The first training method, "from scratch," involves initializing both the Warpformer encoder and classifier from scratch and training them solely on the specified number of samples. The second method, "pretrain," utilizes the best previously learned model parameters for the Warpformer encoder and trains each new classifier on the new task with the given samples while fine-tuning the Warpformer encoder. Additionally, we record the effects of pretrain and from-scratch training of Warpformer under varying epochs, from shallow-tuning (fewer epochs) to deep-tuning (more epochs), for each sample size. This approach allows us to assess the LLM's transferability with greater granularity and better evaluate its transfer capability by comparing it with different training degrees of other models.


\subsection{Backbone Selection and Baseline Contrast}
\label{sec:baseline}
\vspace{-5 pt}
The experiment results for the effect of inference for all the baselines are shown in Section \ref{exp detail}. Figure \ref{fig:baseline} shows that all RNN-based models have similar performance, all of which are better than SEFT. Meanwhile, MTL based on Warpformer significantly outperforms the results of MTL based on various other baselines. Moreover, the box length of Warpformer in the boxplot is relatively narrower, indicating that Warpformer's performance in completing various tasks is relatively consistent.

As Figure \ref{fig:specific baseline} in Section \ref{exp detail}, in different subtasks, the performance of Warpformer as the backbone model for MTL is either better than or at least comparable to the results of other baselines. Meanwhile, it is observed that the Phenotype's performance is slightly inferior. For Phenotype, although the location of boxes of Phenotype is slightly lower compared to some other baselines, the whole whisker is still higher than any other baselines, indicating that Warpformer, when completing Phenotype tasks, although not outstanding overall, does not yield any particularly poor task performance.

For tasks other than Phenotype, the advantage of the Warpformer model is still very apparent. Especially for tasks like Decom and WBM, the worst-performing tasks in these categories are still better than the best-performing tasks of other models. This demonstrates that selecting Warpformer as the unified backbone for MTL is both strong and rational.


\subsection{Comparison of LLM and MTL, STL on inference}
\label{sec:inference}


\begin{figure*}[htbp]
  \centering
    \subfloat[The overall auroc result.]{
    \label{fig:inference}
    \includegraphics[width=0.52\linewidth]{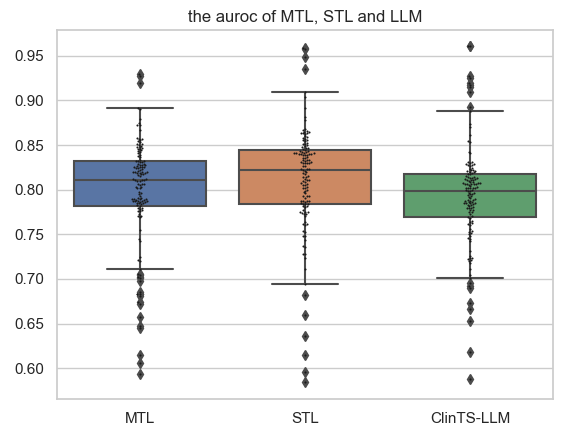}}
    \hspace{1em}
    \subfloat[The specific auroc for each category task.]{
    \label{fig:specific inference}
    \includegraphics[width=0.43\linewidth]{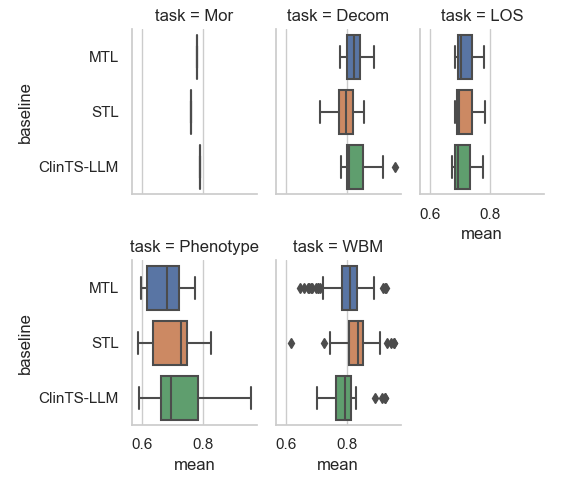}}
    \caption{The behavior of MTL, STL and LLM for inference.}
    \label{fig:all_inference}
\end{figure*}

As shown in Figure \ref{fig:inference}, When LLM serves as a universal decoder, its performance is overall comparable to traditional MTL with hundreds of task heads. For LLM, the box and whisker plot shows very short lengths, indicating the absence of a long tail effect. This suggests that, compared to MTL, LLM has fewer outliers and maintains consistent performance across all tasks without any significant deficiencies.

Furthermore, Figure \ref{fig:specific inference} illustrates that when examining LLM's performance across each task category, it outperforms MTL in all tasks except for WBM tasks. However, due to the substantial number of WBM tasks, which outnumber other task types by several times, they significantly impact the overall model performance. As a result, LLM's performance as a universal decoder may appear slightly inferior overall.

Additionally, it is observed that the overall performance of single-task learning (STL) and multi-task learning (MTL) is very similar, with some tasks even showing better performance under MTL. Thus, employing MTL can significantly enhance efficiency and simplify model management without compromising task performance.

\subsection{LLM Zero-shot Transfer}
\label{sec:zero-transfer}

\subsubsection{Parameter Transfer \& Class Partition Transfer}
\begin{figure*}[t]
  \centering
    \subfloat[Parameter transfer.]{
    \label{fig:combine_win}
    \includegraphics[width=0.48\linewidth]{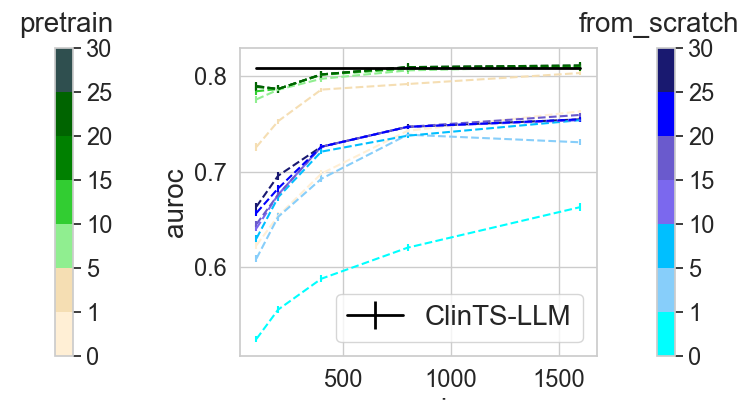}}
    \subfloat[Class partition transfer.]{
    \label{fig:combine_class}
    \includegraphics[width=0.48\linewidth]{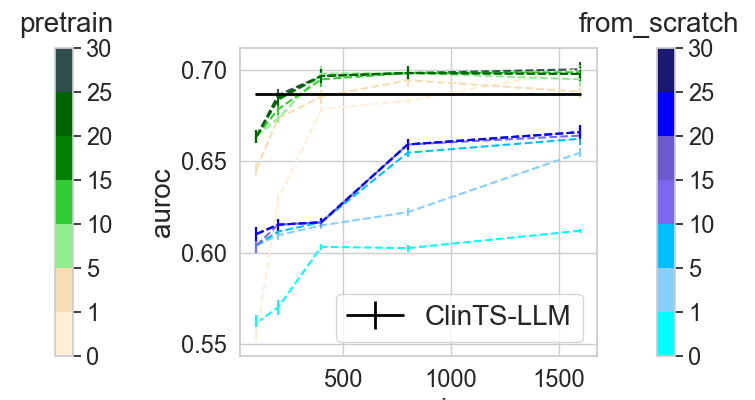}}

    \caption{The zero-shot transfer effect. ClinTS-LLMs is in zero-shot setting and other models are trained with 100, 200, 400, 800, and 1600 samples, shown in the x-axis. The green lines represent the pretrain results, and the blue lines represent the scratch results. The colors transition from light to dark, indicating the process from shallow-tuning to deep-tuning. The specific intensity of the color shown in the color bars on each side of the line chart corresponds to the number of epochs tuned.}
    \label{fig:all_combine}
\end{figure*}

From Figure \ref{fig:all_combine}, we observe that all "from-scratch" Warpformers perform significantly worse than pre-trained ones, indicating that loading parameters from previous tasks has a strong positive effect on completing new tasks. Additionally, when the sample size is very small, deep-tuning significantly outperforms shallow-tuning. However, as the sample size gradually increases, the performance of deep-tuning becomes closer to that of shallow-tuning. Furthermore, the ClinTS-LLM model, even without any sample fine-tuning and using only the model parameters trained on the original task, outperforms all from-scratch results and all pretrain shallow-tuning results.

When compared to pretrain deep-tuning results, for window-size transfer, the LLM can completely surpass them when the pretrain sample-num is small and perform on par with them when the pretrain sample-num is large. For class partition transfer, when the pretrain sample-num is less than 200, the LLM can also completely surpass the pretrain deep-tuning results. However, when the pretrain sample-num is larger, the LLM's performance is lower than that of the pretrain deep-tuning, which is expected and acceptable. After all, when the classification space changes completely, it is quite challenging for the model to achieve successful transfer without any fine-tuning. 

The results above demonstrate the excellent zero-shot transfer capability of LLM as a universal decoder for both parameter transfer and class partition transfer. Figure \ref{fig:combine_win} summarizes the outcomes of all new tasks obtained by altering the window size in-domain and out-domain for the WBM and Decom tasks. Similarly, Figure \ref{fig:combine_class} summarizes the results of all new tasks obtained by changing the task space partition in-domain and out-domain for the LOS task. Detailed experimental results for each specific task can be found in Section \ref{exp detail}, Figure \ref{fig:all_los_wbm_decom}.

\subsubsection{Insights: the in-domain and out-domain transfer}
From Figures \ref{fig:dec_in} and \ref{fig:dec_out}, we observe that when transferring the parameters of Decom to the out-domain of existing parameters, the performance of all models (pretrained and from-scratch, tuned with various sample sizes) significantly decreases. Similarly, from Figures \ref{fig:los_in} and \ref{fig:los_out}, transferring the class partition of LOS to the out-domain of existing class numbers also results in a significant performance drop for all models. However, from Figures \ref{fig:wbm_win_in} and \ref{fig:wbm_win_out}, when transferring the parameters of WBM to the out-domain of existing parameters, the performance of all models (pretrained and from-scratch, tuned with various sample sizes) improves.

Generally, in-domain transfer means that the transferred task falls within the range of the original task space, while out-domain transfer means it falls outside this range. Therefore, in-domain transfer is expected to be less challenging than out-domain transfer. The performance of Decom and LOS transfer tasks supports this notion. However, the result for WBM is unexpected. We believe this may be because WBM indicates whether a certain clinical parameter will be measured over a future period. The out-domain task we set has a longer time window than the existing one. For the WBM task, this indicator tends to stabilize over time. The result of whether the indicator will be measured in the earlier period can sufficiently reflect the outcome of extending the time window further. Therefore, for the WBM task, the out-domain task is more similar to the tasks set during training, resulting in improved performance for all models during transfer.

\subsection{Case Study: Can Few-shot Make up the Weakness for Label Choice Transfer?}
\begin{figure*}[t]
  \centering
    \subfloat[Phenotype zero-shot.]{
    \label{fig:pheno_zero}
    \includegraphics[width=0.48\linewidth]{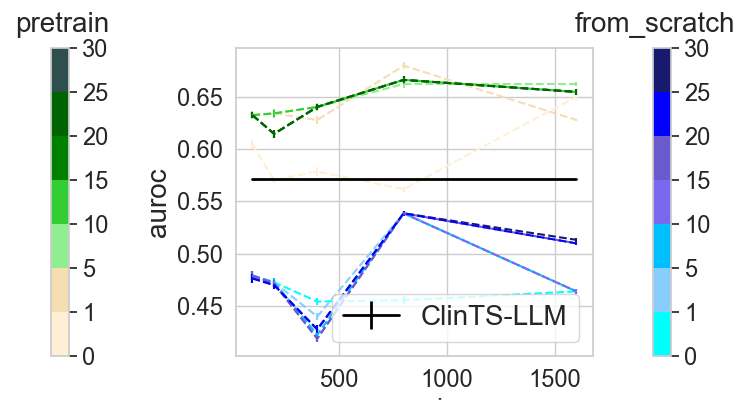}}
    \subfloat[Phenotype few-shot.]{
    \label{fig:pheno_few}
    \includegraphics[width=0.48\linewidth]{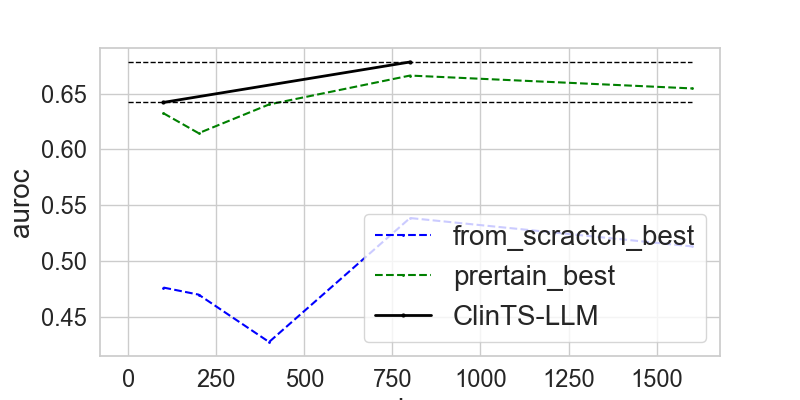}}
    \caption{The label choice transfer effect. All features in subfigure \ref{fig:pheno_zero} are the same as Figure \ref{fig:all_combine}. For subfigure \ref{fig:pheno_few}, we only show the best tune result (selected from all deep-tuning and shallow-tuning) for pretrain and from-scratch model.}
    \label{fig:all_los}
\end{figure*}

From Figure \ref{fig:pheno_zero}, in the zero-shot setting, the performance of LLM is better than all from-scratch models but generally lower than all pretrain models, only slightly surpassing the pretrain results with a small sample size for shallow-tuning. This outcome is not surprising, given that the prediction labels have changed, making the tasks almost entirely different. Therefore, it is understandable that the LLM's performance in zero-shot transfer without any training samples is suboptimal. However, if we provide LLM with a certain number of samples for training and perform few-shot transfer, its capabilities far exceed those of the pretrain models.

As shown in the Figure \ref{fig:pheno_few}, with just 100 samples, the LLM's performance on Phenotype surpasses the best performance of pretrain deep-tuning with 400 samples. Moreover, with 800 samples, the LLM's performance surpasses that of pretrain with any number of samples. Therefore, compared to pretrain, LLM converges faster, achieving the superior results with fewer samples. Additionally, the optimal performance is better, with a higher peak AUROC.

\section{Discussion}
\label{discussion}
This work systematically discusses the limitations of current research on clinical signal tasks. For these works, they focus on only a limited number of task categories and use separate models, or at least separate classifiers to train these tasks. Based on this, proposes a comprehensive clinical task framework. Additionally, We have developed a unified model that employs LLM as a universal decoder to manage various tasks within our framework. Through detailed experiments, we have demonstrated that our model, utilizing a single universal decoder (LLM), can match the performance of traditional multi-task learning with hundreds of task-specific heads when inferring existing tasks. Furthermore, when transferring to entirely new tasks, our model exhibits strong zero-shot and few-shot capabilities.

Nonetheless, our work has certain limitations. Although we proposed multiple perspectives to expand a comprehensive clinical task framework from core clinical tasks, this framework can be further refined by incorporating additional core tasks (such as Intervene, Final Acuity, etc.). Additionally, our model's zero-shot performance in label choice transfer tasks is suboptimal and still requires few-shot training with a small number of samples to achieve optimal results. Future work could focus on improving the model to enhance its performance in all types of zero-shot transfer. Lastly, our research is limited to the clinical domain and has not explored whether LLM has the potential to serve as a universal decoder in other fields, which warrants further investigation.


\bibliographystyle{elsarticle-num}
\bibliography{refs}


\appendix

\section{Appendix / supplemental material}

\subsection{The Basis for Dividing the Bataset according to the Standard}
\label{basis dividing dataset}

When dividing the dataset, we followed the approach of work~\cite{mcdermott2021comprehensive}, using a time-based method to partition all samples. The greatest advantage of this approach is that it aligns with our general understanding and common demands for time series tasks. We hope to use existing data to provide guidance for future clinical measures. 

However, previous work~\cite{harutyunyan2019multitask,zhang2023warpformer} chose to divide the dataset based on complete patients, considering that dividing the dataset based on the time axis may result in data leakage. For example, the state of Mor in the same patient during one HADM remains unchanged, so the label information may be leaked to the later samples on the time axis. This data leakage problem is essentially since there may be some overlap in the time axis that affect the same task in the training set, validation set, and test set. 

Therefore, in this work, although we still use a time-based approach, we still ensure that we use HADM (the maximum length that a task can span) as the unit of division. 

\subsection{The Differences between our clinical Tasks and Those of other Related Works}

\label{differences in clinical tasks}
\emph{\textbf{In-hospital Mortality (MOR).}}
The goal of this task is to predict whether a patient will die during their current hospital admission period (HADM). The input for this task is derived from a 24-hour sliding window of data. In the label, a value of 1 indicates that the patient has died by the end of the current hospital admission, while a value of 0 indicates that the patient has not died. This approach to constructing the task differs from previous work, as it does not limit the input to the first 48 hours~\cite{harutyunyan2019multitask} or all monitoring data prior to the current time point~\cite{mcdermott2021comprehensive}. Instead, the input is segmented using a sliding window approach, allowing the model to accurately predict the patient's condition at any stage of their illness. This design is also consistent with our goals to design a data framework where each individual input corresponds to all labels.

\emph{\textbf{Decompensation (Decom).}}
The goal of this task is to predict whether a patient will die within a certain period of time in the future. In this task, our input data is consistent with the previous tasks and is derived from a 24-hour sliding window. To determine the label value corresponding to each input, we need to define a future window, as in some works~\cite{harutyunyan2019multitask,mcdermott2021comprehensive,zhang2023warpformer}, to determine whether the patient will die (label value is set to 1) or survive (label value is set to 0) within the future window. In some works~\cite{harutyunyan2019multitask,zhang2023warpformer}, the future window is directly defined as the next 24 hours. However, more works~\cite{wang2020mimic} did not define the Decom task and pointed out that the nature of the Decom task comes from the Mor task, which is equivalent to the Mor task within a certain period of time. Therefore, our work is compatible with the 24-hour future window proposed by those previous work~\cite{harutyunyan2019multitask,zhang2023warpformer}, while also taking into account the nature of the Decom task proposed by [extraction]: we designed a series of future windows from 3h to 120h, with a 3h increment, and each future window corresponds to a separate Decom task.

\emph{\textbf{Length Of Stay (LOS).}}
The goal of this task is to predict the total length of stay in icu for a patient. In this task, our input data is consistent with the previous tasks and is derived from a 24-hour sliding window. Referring to some works~\cite{mcdermott2021comprehensive} defined the length of stay as a binary classification task, with a length of stay less than 3 as one class and the rest as another class. Some other works~\cite{zhang2023warpformer} defined the length of stay as a 9-classification task, with categories of 1~7 days, < 2 weeks, > 2 weeks, and many other works have proposed different classification criteria. Considering that all previous works have transformed a regression task into a multi-classification task, there are naturally different standards. Therefore, for the LOS task, we did not adopt a unique partition method like previous works, but instead defined a partition method set, which includes multiple different partition methods for the LOS task. The central idea behind the specific partition method set is to make the sample quantities of different categories as balanced as possible for each partition. Therefore, for the n-classification partition method in the partition method set, the construction method is as follows: sort all samples in order of their specific length of stay from small to large, set n equidistant points based on the sample quantity, and then find the integer values of the length of stay corresponding to the n equidistant points as the boundaries for the classifications under that partition method.

\emph{\textbf{Phenotype.}}
The goal of this task is to determine which diseases a patient has in their current HADM. In this task, our input is unified with previous tasks and is derived from a 24-hour sliding window. Referring to some works~\cite{harutyunyan2019multitask}, we also realize that there are numerous phenotypes in MIMIC-III, so we select the most representative ones from all phenotypes. The original information related to phenotypes in the MIMIC-III dataset is reflected in the DIAGNOSIS attribute column. However, each DIAGNOSIS represents all diseases diagnosed after a patient's HADM ends. Therefore, if we want to select some of the more frequent diseases, we need to segment them into individual phenotypes. At the same time, after selecting the more frequent phenotypes, since these phenotypes may have errors or missing letters or abbreviations in the process of recording data, multiple independent phenotypes with small differences may point to the same type of disease, and therefore need to be merged, such as PNEUMO and PNEUMONIA, which should be ultimately merged into PNEUMONIA. In addition, some diseases are too general and can be composed of other phenotypes. In this case, we need to use the decomposed phenotypes to replace the original general phenotype, such as using GASTROINTESTINAL BLEED and UPPER GI BLEED to replace BLEED (because BLEED is too general).

\emph{\textbf{Next Timepoint Will Be Measured (WBM).}}
The goal of this task is to predict which indicators will be measured for the patient in a future period. In this task, the input is consistent with the previous tasks, and is derived from a 24-hour sliding window. We chose the 56 indicators, which are the same with some works~\cite{zhang2023warpformer}, as the selected indicators. However, there is a significant difference in the selection of the future window in different works~\cite{mcdermott2021comprehensive} choosing a 1-hour future window, while the other works~\cite{zhang2023warpformer} chooses a 24-hour future window. Since WBM will lose its ability to monitor the patient's vital signs if the future window is too long, we have set a series of future windows considering all the previous works, ranging from 0.5h to 6h, with an increment of 0.5h, and each future window corresponds to a separate WBM task.

\subsection{Experiment Details}
\label{exp detail}

\subsubsection{Experimental Settings}
\label{sec:exp_setting}
We implement 1 transformer block with 2 heads for Warpformer. The rank of LoRA is 32. We optimize the model with AdamW, with a 3e-4 learning rate and 240 batch size, distributing on 4 A6000 cards with Fully Sharded Data Parallel in BF16 data type, without any learning rate scheduler. 

\subsubsection{Evaluation Metrics}
\label{sec:evaulate_metric}
Due to the severe label distribution imbalance in clinical tasks, evaluating model performance using accuracy is likely to overestimate overall performance and prevent fair comparison of different methods. Therefore, in line with previous research, we used the area under the receiver operating characteristic curve (AUROC) to evaluate the performance of each task in the MIMIC-III dataset. Additionally, due to the large number of tasks in our set, many of which have not been previously considered in other works, the area under the precision-recall curve (AUPRC) is not a wise choice as an evaluation metric, as it performs poorly across all methods for many of these tasks. As a result, We present all evaluation metrics as AUROC. For multi-class or multi-label tasks, we calculated the AUROC score for each class (label) and then computed the average as the final score.

\subsection{Experiment results}
\label{ap exp result}

\begin{figure*}[htbp]
  \centering
    \subfloat[The overall auroc result.]{
    \label{fig:baseline}
    \includegraphics[width=0.78\linewidth]{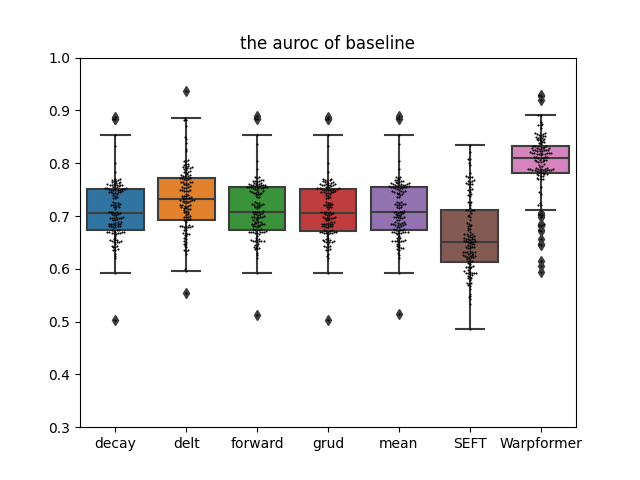}}

    \subfloat[The specific auroc for each category task.]{
    \label{fig:specific baseline}
    \includegraphics[width=0.78\linewidth]{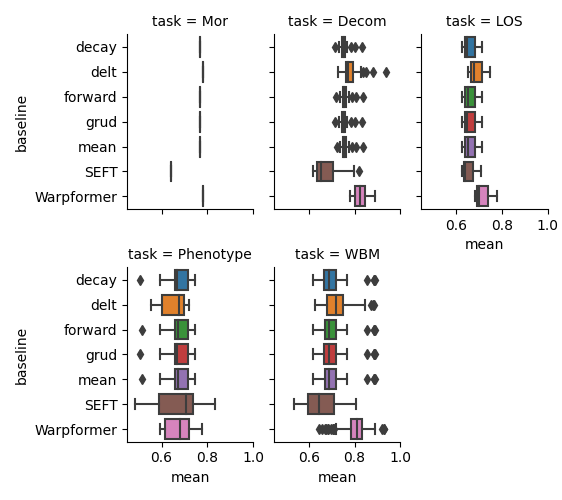}}
    \caption{The behavior of all the baseline for MTL.}
    \label{fig:all_baseline}
\end{figure*}

\begin{figure*}[htbp]
  \centering
  
    \subfloat[Decom window in-domain.]{
    \label{fig:dec_in}
    \includegraphics[width=0.48\linewidth]{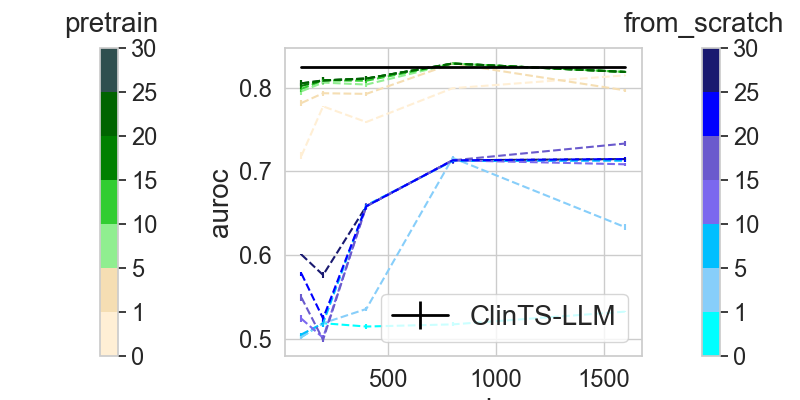}}
    \subfloat[Decom window out-domain.]{
    \label{fig:dec_out}
    \includegraphics[width=0.48\linewidth]{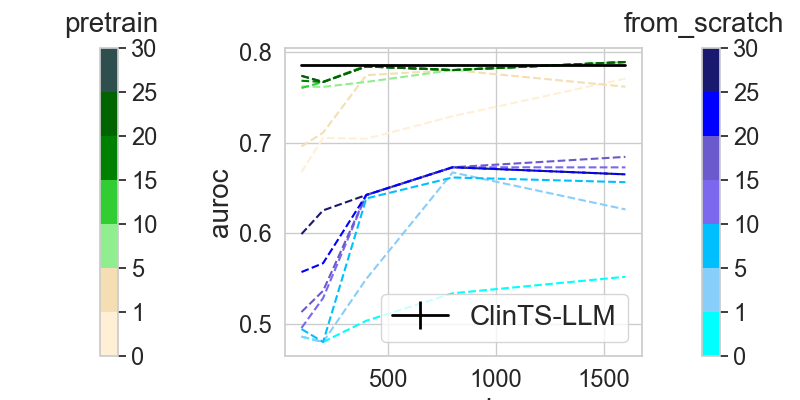}}
    
    \subfloat[WBM window in-domain.]{
    \label{fig:wbm_win_in}
    \includegraphics[width=0.48\linewidth]{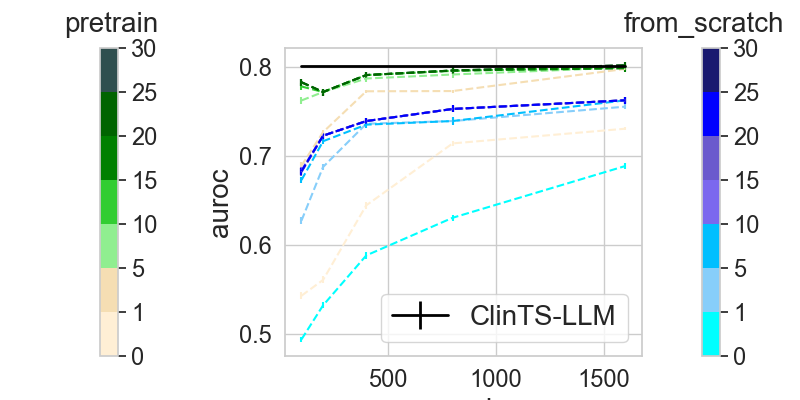}}
    \subfloat[WBM window out-domain.]{
    \label{fig:wbm_win_out}
    \includegraphics[width=0.48\linewidth]{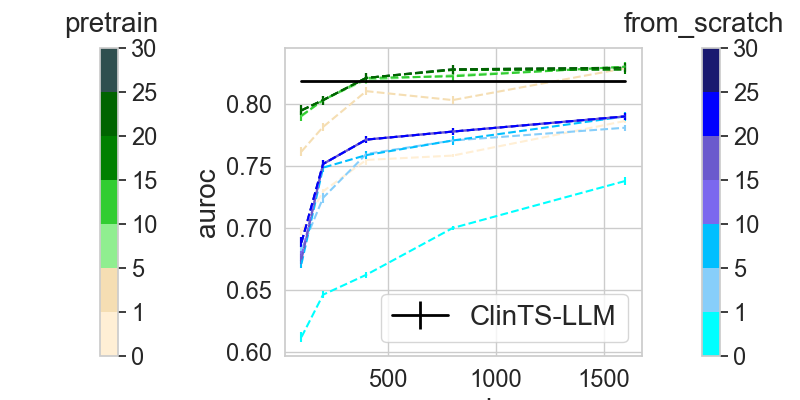}}

    \subfloat[LOS window in-domain.]{
    \label{fig:los_in}
    \includegraphics[width=0.48\linewidth]{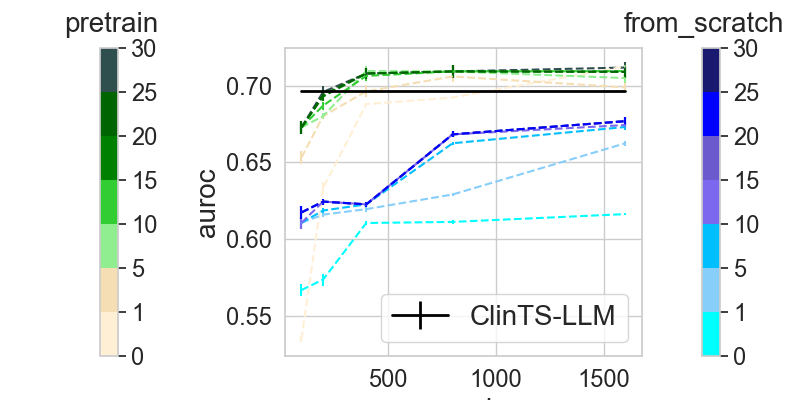}}
    \subfloat[LOS window out-domain.]{
    \label{fig:los_out}
    \includegraphics[width=0.48\linewidth]{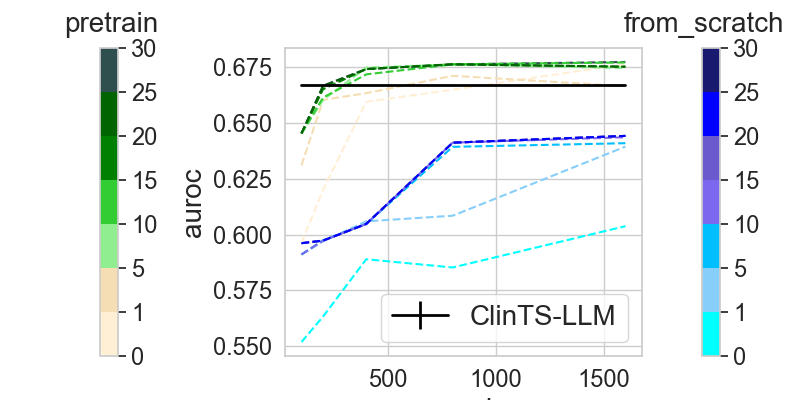}}
    
    \caption{The detail of LOS, WBM, and Decom transfer tasks of class partition transfer and parameter transfer, concluding in-domain and out-domain transfer results.}
    \label{fig:all_los_wbm_decom}
\end{figure*}

\centering
\begin{longtable}{lcccc}

\caption{The statics details for all the clinical tasks in experiment. } 
\label{tab:all_task} \\

\hline \textbf{Task.seq}  & \textbf{Category}  &  \textbf{Transfer type} & \textbf{Window size} & \textbf{Label} 
\\
\midrule

\endfirsthead

\hline \textbf{Task.seq}  & \textbf{Category}  &  \textbf{Transfer type} & \textbf{Window size} & \textbf{Label} \\
\midrule
\endhead

\hline
\endfoot

1  & Mor  &  Train task  &  &   \\  
2  & Decom  &  Train task  &  3h &   \\
3  & Decom  &  Train task  &  6h &   \\
4  & Decom  &  Train task  &  9h &   \\
5  & Decom  &  Train task  &  12h &   \\
6  & Decom  &  Train task  &  15h &   \\
7  & Decom  &  Train task  &  18h &   \\
8  & Decom  &  Train task  &  21h &   \\
9  & Decom  &  Train task  &  24h &   \\
10  & Decom  &  Train task  &  27h &   \\
11  & Decom  &  Train task  &  33h &   \\
12  & Decom  &  Train task  &  36h &   \\
13  & Decom  &  Train task  &  39h &   \\
14  & Decom  &  Train task  &  42h &   \\
15  & Decom  &  Train task  &  48h &   \\
16  & Decom  &  Train task  &  51h &   \\
17  & Decom  &  Train task  &  54h &   \\
18  & Decom  &  Train task  &  57h &   \\
19  & Decom  &  Train task  &  60h &   \\
20  & Decom  &  Train task  &  63h &   \\
21  & Decom  &  Train task  &  66h &   \\
22  & Decom  &  Train task  &  69h &   \\
23  & Decom  &  Train task  &  72h &   \\
24  & Decom  &  Train task  &  78h &   \\
25  & Decom  &  Train task  &  81h &   \\
26  & Decom  &  Train task  &  84h &   \\
27  & Decom  &  Train task  &  87h &   \\
28  & Decom  &  Train task  &  93h &   \\
29  & Decom  &  Train task  &  96h &   \\
30  & Decom  &  Train task  &  99h &   \\
31  & Decom  &  Train task  &  102h &   \\
32  & Decom  &  Train task  &  105h &   \\
33  & Decom  &  Train task  &  108h &   \\
34  & Decom  &  Train task  &  111h &   \\
35  & Decom  &  Train task  &  114h &   \\
36  & Decom  &  Train task  &  117h &   \\
37  & LOS  &  Train task  &   & 0-5,more\\
38  & LOS  &  Train task  &   & 0-2,2-5,5-12,more\\
39  & LOS  &  Train task  &   & 0-1,1-3,3-5,5-9,9-16,more\\
40  & Phenotype  &  Train task  &   & ORONARY ARTERY DISEASE\\
41  & Phenotype  &  Train task  &   & PNEUMONIA\\
42  & Phenotype  &  Train task  &   & ITIS\\
43  & Phenotype  &  Train task  &   & SEPSIS\\
44  & Phenotype  &  Train task  &   & HEART FAILURE\\
45  & Phenotype  &  Train task  &   & CHEST PAIN\\
46  & Phenotype  &  Train task  &   & MYOCARDIAL INFARCTION\\
47  & Phenotype  &  Train task  &   & GASTROINTESTINAL BLEED\\
48  & Phenotype  &  Train task  &   & FEVER\\
49  & WBM  &  Train task  & 0.5h  & FiO2\\
50  & WBM  &  Train task  & 0.5h  & Glucose\\
51  & WBM  &  Train task  & 0.5h  & Sodium\\
52  & WBM  &  Train task  & 0.5h  & Potassium\\
53  & WBM  &  Train task  & 0.5h  & Magnesium\\
54  & WBM  &  Train task  & 0.5h  & Hct\\
55  & WBM  &  Train task  & 0.5h  & Chloride\\
56  & WBM  &  Train task  & 0.5h  & pH Blood\\
57  & WBM  &  Train task  & 0.5h  & Total CO2\\
58  & WBM  &  Train task  & 0.5h  & Base Excess\\
59  & WBM  &  Train task  & 1h  & FiO2\\
60  & WBM  &  Train task  & 1h  & Glucose\\
61  & WBM  &  Train task  & 1h  & Sodium\\
62  & WBM  &  Train task  & 1h  & Potassium\\
63  & WBM  &  Train task  & 1h  & Magnesium\\
64  & WBM  &  Train task  & 1h  & Hct\\
65  & WBM  &  Train task  & 1h  & Chloride\\
66  & WBM  &  Train task  & 1h  & pH Blood\\
67  & WBM  &  Train task  & 1h  & Total CO2\\
68  & WBM  &  Train task  & 1h  & Base Excess\\
69  & WBM  &  Train task  & 1.5h  & FiO2\\
70  & WBM  &  Train task  & 1.5h  & Glucose\\
71  & WBM  &  Train task  & 1.5h  & Sodium\\
72  & WBM  &  Train task  & 1.5h  & Potassium\\
73  & WBM  &  Train task  & 1.5h  & Magnesium\\
74  & WBM  &  Train task  & 1.5h  & Hct\\
75  & WBM  &  Train task  & 1.5h  & Chloride\\
76  & WBM  &  Train task  & 1.5h  & pH Blood\\
77  & WBM  &  Train task  & 1.5h  & Total CO2\\
78  & WBM  &  Train task  & 1.5h  & Base Excess\\
79  & WBM  &  Train task  & 2h  & FiO2\\
80  & WBM  &  Train task  & 2h  & Glucose\\
81  & WBM  &  Train task  & 2h  & Sodium\\
82  & WBM  &  Train task  & 2h  & Potassium\\
83  & WBM  &  Train task  & 2h  & Magnesium\\
84  & WBM  &  Train task  & 2h  & Hct\\
85  & WBM  &  Train task  & 2h  & Chloride\\
86  & WBM  &  Train task  & 2h  & pH Blood\\
87  & WBM  &  Train task  & 2h  & Total CO2\\
88  & WBM  &  Train task  & 2h  & Base Excess\\
89  & WBM  &  Train task  & 2.5h  & FiO2\\
90  & WBM  &  Train task  & 2.5h  & Glucose\\
91  & WBM  &  Train task  & 2.5h  & Sodium\\
92  & WBM  &  Train task  & 2.5h  & Potassium\\
93  & WBM  &  Train task  & 2.5h  & Magnesium\\
94  & WBM  &  Train task  & 2.5h  & Hct\\
95  & WBM  &  Train task  & 2.5h  & Chloride\\
96  & WBM  &  Train task  & 2.5h  & pH Blood\\
97  & WBM  &  Train task  & 2.5h  & Total CO2\\
98  & WBM  &  Train task  & 2.5h  & Base Excess\\
99  & WBM  &  Train task  & 4h  & FiO2\\
100  & WBM  &  Train task  & 4h  & Glucose\\
101  & WBM  &  Train task  & 4h  & Sodium\\
102  & WBM  &  Train task  & 4h  & Potassium\\
103  & WBM  &  Train task  & 4h  & Magnesium\\
104  & WBM  &  Train task  & 4h  & Hct\\
105  & WBM  &  Train task  & 4h  & Chloride\\
106  & WBM  &  Train task  & 4h  & pH Blood\\
107  & WBM  &  Train task  & 4h  & Total CO2\\
108  & WBM  &  Train task  & 4h  & Base Excess\\
109  & WBM  &  Train task  & 4.5h  & FiO2\\
110  & WBM  &  Train task  & 4.5h  & Glucose\\
111  & WBM  &  Train task  & 4.5h  & Sodium\\
112  & WBM  &  Train task  & 4.5h  & Potassium\\
113  & WBM  &  Train task  & 4.5h  & Magnesium\\
114  & WBM  &  Train task  & 4.5h  & Hct\\
115  & WBM  &  Train task  & 4.5h  & Chloride\\
116  & WBM  &  Train task  & 4.5h  & pH Blood\\
117  & WBM  &  Train task  & 4.5h  & Total CO2\\
118  & WBM  &  Train task  & 4.5h  & Base Excess\\
119  & WBM  &  Train task  & 5h  & FiO2\\
120  & WBM  &  Train task  & 5h  & Glucose\\
121  & WBM  &  Train task  & 5h  & Sodium\\
122  & WBM  &  Train task  & 5h  & Potassium\\
123  & WBM  &  Train task  & 5h  & Magnesium\\
124  & WBM  &  Train task  & 5h  & Hct\\
125  & WBM  &  Train task  & 5h  & Chloride\\
126  & WBM  &  Train task  & 5h  & pH Blood\\
127  & WBM  &  Train task  & 5h  & Total CO2\\
128  & WBM  &  Train task  & 5h  & Base Excess\\
129  & WBM  &  Train task  & 5.5h  & FiO2\\
130  & WBM  &  Train task  & 5.5h  & Glucose\\
131  & WBM  &  Train task  & 5.5h  & Sodium\\
132  & WBM  &  Train task  & 5.5h  & Potassium\\
133  & WBM  &  Train task  & 5.5h  & Magnesium\\
134  & WBM  &  Train task  & 5.5h  & Hct\\
135  & WBM  &  Train task  & 5.5h  & Chloride\\
136  & WBM  &  Train task  & 5.5h  & pH Blood\\
137  & WBM  &  Train task  & 5.5h  & Total CO2\\
138  & WBM  &  Train task  & 5.5h  & Base Excess\\
139  & Decom  &  Window size  &  30h &   \\
140  & Decom  &  Window size  &  45h &   \\
141  & Decom  &  Window size  &  75h &   \\
142  & Decom  &  Window size  &  90h &   \\
143  & Decom  &  Window size  &  120h &   \\
144  & LOS  &  Class partition  &   & 0-3,3-9,more\\
145  & LOS  &  Class partition  &   & 0-1,1-3,3-7,7-14,more\\
146  & LOS  &  Class partition  &   & 0-1,1-2,2-4,4-6,6-10,10-17,more\\
147 & Phenotype  &  Label chocie  &   & AORTIC STENOSIS \\
148 & Phenotype  &  Label chocie  &   & RENAL FAILURE \\
149 & Phenotype  &  Label chocie  &   & UPPER GI BLEED \\
150 & Phenotype  &  Label chocie  &   & HYPOT \\
151 & Phenotype  &  Label chocie  &   & ALTERED MENTAL STATUS \\
152  & WBM  &  Window size  & 3h  & FiO2\\
153  & WBM  &  Window size  & 3h  & Glucose\\
154  & WBM  &  Window size  & 3h  & Sodium\\
155  & WBM  &  Window size  & 3h  & Potassium\\
156  & WBM  &  Window size  & 3h  & Magnesium\\
157  & WBM  &  Window size  & 3.5h  & FiO2\\
158  & WBM  &  Window size  & 3.5h  & Glucose\\
159  & WBM  &  Window size  & 3.5h  & Sodium\\
160  & WBM  &  Window size  & 3.5h  & Potassium\\
161  & WBM  &  Window size  & 3.5h  & Magnesium\\
162  & WBM  &  Window size  & 6h  & FiO2\\
163  & WBM  &  Window size  & 6h  & Glucose\\
164  & WBM  &  Window size  & 6h  & Sodium\\
165  & WBM  &  Window size  & 6h  & Potassium\\
166  & WBM  &  Window size  & 6h  & Magnesium\\
\bottomrule
\end{longtable}

\end{document}